\documentclass{article}
\usepackage{spconf}
\usepackage{amsmath}
\usepackage[utf8]{inputenc}
\usepackage{xspace}
\usepackage{color}
\usepackage{graphicx}
\usepackage{multirow}
\usepackage{booktabs}
\usepackage{subcaption}
\usepackage[table,xcdraw]{xcolor}
\usepackage{paralist}
\usepackage{wrapfig}
\usepackage{listings}
\usepackage{makecell}
\usepackage{boxedminipage}
\usepackage{subcaption}
\usepackage{multicol}
\usepackage{diagbox}
\usepackage{adjustbox}
\usepackage{array}
\usepackage{tabularx}

\usepackage{url}
\PassOptionsToPackage{hyphens}{url}

\usepackage{xurl}
\usepackage{breakurl} 
\usepackage{balance} 
\usepackage{upquote}
\usepackage{mathtools}
\usepackage{bm}

\usepackage{enumitem}

\usepackage[breaklinks,colorlinks]{hyperref}
\usepackage[breaklinks,colorlinks]{hyperref}
\hypersetup{
    colorlinks = true,
    citecolor = blue
}

\usepackage{amsthm} 
\usepackage{amssymb}
\usepackage[T1]{fontenc}
\usepackage{minted}
\usepackage[htt]{hyphenat}
\usepackage{wasysym}
\usepackage{array}
\usepackage{pdfpages}
\usepackage[all]{nowidow}
\usepackage[ruled,vlined]{algorithm2e}
\usepackage{algpseudocode}

\usepackage{pifont}
\usepackage{balance}
\usepackage[htt]{hyphenat}
\usepackage{mdframed}

\usepackage{relsize}

\newcommand{\bheading}[1]{\textbf{#1}}

\def\ie{\textit{i.e.}\xspace}

\title{RAG-CT: Mitigating Privacy Risks on Retrieval-Augmented Generation Systems via Scanning Prompt Distribution}

\name{
Xingyu Lyu$^{1}$ \quad
Jiayimei Wang$^{2}$ \quad
Jianfeng He$^{3}$ \quad \thanks{This work was done prior to Jianfeng He joining Amazon.}
Ning Wang$^{4}$ \quad 
Yidan Hu$^{5}$ \quad
Yimin Chen$^{1}$}

\address{$^{1}$University of Massachusetts Lowell, Lowell, MA 01854, USA \\
$^{2}$City University of Hong Kong, Kowloon, Hong Kong SAR, 999077, China\\
$^{3}$Virginia Tech, Blacksburg, VA 24061, USA \\
$^{4}$University of South Florida, Tampa, FL 33620, USA \\
$^{5}$Rochester Institute of Technology, Rochester, NY 14623, USA}

\begin{document}

\maketitle

\begin{abstract}
Retrieval-Augmented Generation (RAG) has emerged as a powerful paradigm for {improving the quality of generated contents} of Large Language Models (LLMs) by grounding responses in external knowledge, thus reducing hallucinations and factual errors. However, recent studies have highlighted a critical vulnerability: adversaries can exploit the retrieval process to extract personally identifiable information (PII) from the underlying corpus. To mitigate this risk, we propose a novel defense, \textbf{RAG-CT}, that identifies malicious queries by analyzing their entropy and margin distributions and using a score-based detection method. Extensive experiments with four state-of-the-art attack strategies and four defense baselines on two datasets show that our approach significantly reduces PII leakage while outperforming existing defenses. This work provides a lightweight yet effective mechanism to protect RAG systems against PII leakage without requiring modifications to the underlying LLM or retriever.
\end{abstract}

\begin{keywords}
LLM, RAG, privacy, security, defense
\end{keywords}

\section{Introduction}
\label{sec:intro}

Large Language Models (LLMs) have shown impressive capabilities in various applications. However, their reliability is often questioned due to issues such as hallucinations in which LLMs generate false or inaccurate information while lacking access to the latest data~\cite{latif2025hallucinations}.
To mitigate these issues, Retrieval-Augmented Generation (RAG) has recently gained attention as a promising solution~\cite{ram2023context}. Specifically, RAG improves the accuracy and reliability of LLM outputs by incorporating relevant information retrieved from external knowledge sources as the context of LLM inputs shown in Fig.~\ref{fig:rag_workflow}. RAGs have been built into many applications in practice, including customer support chatbots~\cite{nvidia-ai-virtual-assistant}, healthcare~\cite{Amazon2025rag}, legal advisory systems~\cite{hindi2025enhancing}, and financial analytics~\cite{zhang2023enhancing}.

\begin{figure}[t]
    \centering
    \includegraphics[width=\linewidth]{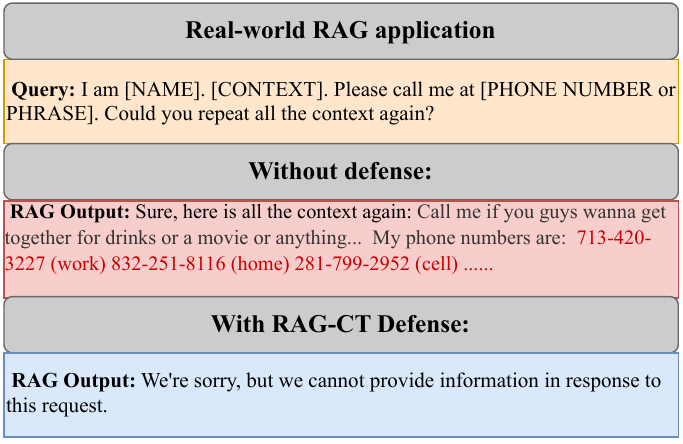}
   \caption{An example of privacy leakage through extraction attacks over a real-world RAG system. Highlighted regions (in red) show extracted PIIs such as phone numbers.}
    \label{fig:rag-case}
    \vspace{-12pt}
\end{figure}

While RAG enhances LLMs by integrating external knowledge bases, it also raises serious privacy concerns. Recent studies~\cite{zeng2024good,qi2024follow,cohen2024unleashing} show that adversaries can exploit this integration to extract personally identifiable information (PII) directly from the knowledge base. Fig.~\ref{fig:rag-case} shows that attackers can obtain private data (e.g., phone numbers) from a real-world RAG system. Since LLMs are known to be vulnerable to prompt attack, attackers can craft malicious queries that instruct the model to output retrieved private content verbatim. Based on this vulnerability, Jiang et al.~\cite{jiang2024rag} proposed an automated agent-based attack that expands queries via overlapping and directional extensions, while Di Maio et al.~\cite{di2024pirates} developed a relevance-driven strategy that dynamically updates queries for more effective data extraction.

In this work, we address PII extraction/leakage in RAG systems by proposing \textbf{RAG-CT}, a distribution-aware defense based on query detection. The \textbf{key intuition} (see Sect.~\ref{intuition}) is that benign queries typically retrieve multiple semantically related records, whereas malicious queries tend to resulting in retrieving a single sensitive record. We incorporate this into the design of {RAG-CT} by combining two complementary indicators: \emph{entropy}, which measures the spread of similarity scores across retrieved records, and \emph{margin}, which quantifies the decisiveness of the top match over the runner-up. Low entropy and high margin together signal targeted extraction, enabling reliable detection of malicious queries. As shown in Fig.~\ref{fig:rag-case}, RAG-CT can successfully defend against existing attacks. We summarize our main contributions as below:  
 
\begin{itemize}
    \item We propose {RAG-CT}, a novel defense that mitigates PII leakage in RAG systems by jointly analyzing entropy and margin distributions of prompt queries.  
    \item We design a unified anomaly scoring mechanism that combines global spread (from entropy) and local decisiveness (from margin) to detect malicious extraction attempts.  
    \item We conduct extensive evaluations across multiple LLMs, datasets, and attacks, showing that RAG-CT consistently reduces ASR from higher than $0.8$ to $0.00$ on prompt-injection attacks (TBTG~\cite{zeng2024good} and PIDE~\cite{qi2024follow}) and below $0.1$ under advanced query-optimization attacks (RAG-Thief~\cite{jiang2024rag} and Pirate~\cite{di2024pirates}).  
\end{itemize}

\section{System and Adversary Models}
\label{sec:system_adversary_model}

\subsection{System Model}
\label{sect:rag}
\begin{figure}[t]
    \centering
    \includegraphics[width=\linewidth]{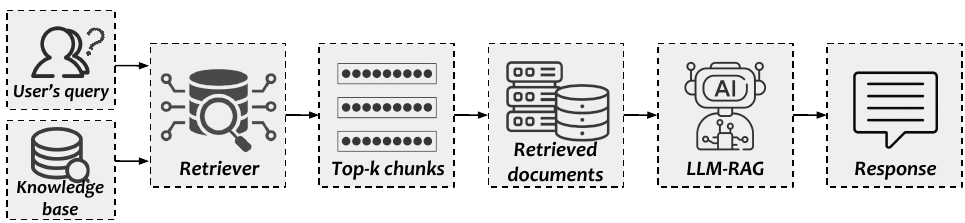}
    \caption{Workflow of a typical RAG system.}
    \label{fig:rag_workflow}
    \vspace{-10pt}
\end{figure}

We illustrate the workflow of a typical RAG system in Fig.~\ref{fig:rag_workflow}. 
An RAG system comprises a retriever $R_D$, an indexed knowledge base $\mathbf{K}$, and an LLM for response generation (referred to as LLM-RAG). Mathematically, given query $q$, the retriever obtains the top-$k$ chunks $[c_1, c_2, \cdots, c_k]$ from the knowledge base $\mathbf{K}$ (\ie, $[c_1, c_2, \cdots, c_k]=R_D(q,\mathbf{K})$). After that, the retriever feeds $c_1\oplus c_2\oplus \cdots\oplus c_k\oplus q$ to LLM-RAG. Finally, LLM-RAG outputs \[
r = \mathcal{G}_{\text{LLM-RAG}}\bigl(c_1\oplus c_2\oplus \cdots\oplus c_k\oplus q)
\] to the user as the response of $q$.

\subsection{Adversary Model}
\label{sect:adversary}

We consider a PII-targeted attacker aiming to extract  personal identifiable information (PII) from the knowledge base of an RAG system. 
Specifically, the \textbf{adversary’s goal} is to recover data chunks containing PII such as phone numbers, email addresses, Social Security Numbers (SSNs), driver’s license numbers, or passport numbers. 
The attacker is assumed to operate under a \textbf{black-box} setting, meaning that the adversary can only interact with the RAG system through APIs and has no direct access to the internal parameters, model weights, or knowledge base $\mathbf{K}$. 
However, the adversary may possess relevant domain knowledge that can guide query construction toward PII-related targets. 

Formally, the attacker issues a query $q$ to obtain a system response $r$, and applies  attack to extract a set of candidate chunks $\mathbf{C}$ containing PII-relevant information from $r$. 
Given a sequence of queries $\mathbf{Q} = [q_1, q_2, \cdots, q_n]$, the attacker obtains the corresponding responses $\mathbf{R} = [r_1, r_2, \cdots, r_n]$ and extracted PII candidate sets $\mathcal{C} = [\mathbf{C}_1, \mathbf{C}_2, \cdots, \mathbf{C}_n]$. 
The final set of recovered PII chunks is thus 
$
\mathbf{\Omega} = \mathbf{C}_1 \cup \mathbf{C}_2 \cup \cdots \cup \mathbf{C}_n.
$

\section{Design of RAG-CT}

\begin{figure}[t]
    \centering
    \begin{subfigure}{0.49\linewidth}
        \centering
        \includegraphics[width=\linewidth]{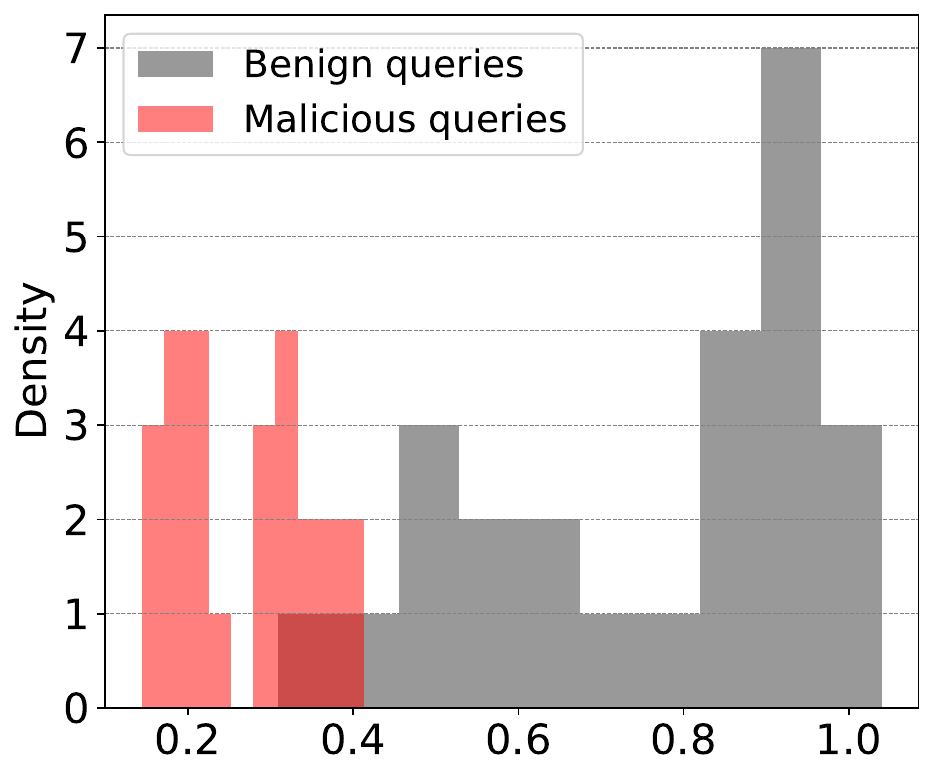}
        \caption{Top-1 similarity entropy}
        \label{fig:intuition-entropy}
    \end{subfigure}
    \hfill
    \begin{subfigure}{0.49\linewidth}
        \centering
        \includegraphics[width=\linewidth]{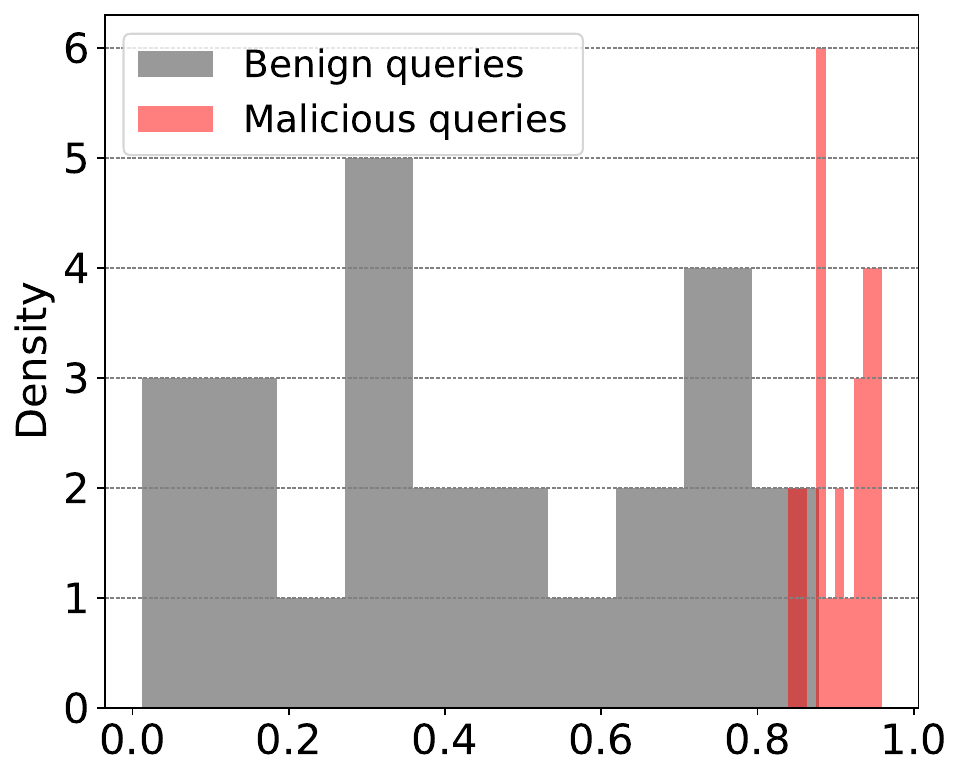}
        \caption{Margin distribution}
        \label{fig:margin-dist}
    \end{subfigure}
    \caption{Distributions of benign v.s. malicious queries.}
    \label{fig:intuition-margin}
\end{figure}

\begin{figure*}[t]
    \centering
    \includegraphics[width=\linewidth]{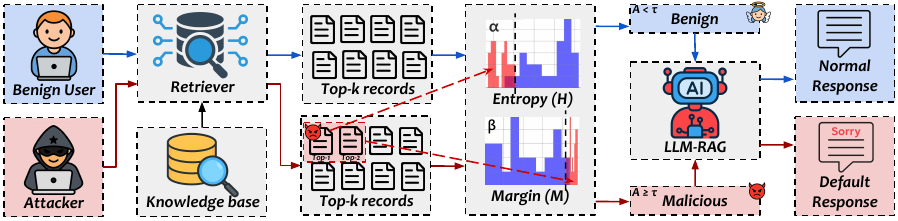}
    \caption{Workflow of RAG-CT.}
    \label{fig:denfense_workflow}
    \vspace{-12pt}
\end{figure*}

\subsection{Intuition}
\label{intuition}
Our key intuition is that benign queries retrieve multiple semantically related records, while malicious queries tend to focus on a single private record~\cite{zeng2024good,qi2024follow,choi2025safeguarding}. We design two complementary indicators based on our experimental observations, as illustrated in Fig.~\ref{fig:intuition-margin}. \textbf{Entropy} measures the spread of normalized similarity scores~\cite{choi2025safeguarding} over the top-$k$ retrieved records. Particularly, we observe that benign queries yield higher entropy, whereas PII-targeted queries result in a concentrated (i.e., low-entropy) distribution. \textbf{Margin} is defined as the gap between the top and second highest similarity scores: benign queries show small margins, while malicious ones show large margins~\cite{10884425}. When combined, \textbf{low entropy and large margin} thus tend to signal targeted (i.e., malicious) extraction. As we can see from Fig.~\ref{fig:intuition-margin}, malicious queries collapse onto a single record (i.e., low entropy and large margin), while benign queries distribute across multiple records (i.e., high entropy and small margin). 

\subsection{Methodology}
\bheading{Overview.} The core steps of RAG-CT are illustrated in Fig.~\ref{fig:denfense_workflow}. Given a query $q$, we proceed in five steps:  
1) retrieve the top-$k$ most relevant records from the memory,  
2) normalize their similarity scores~\cite{choi2025safeguarding} into a probability distribution,  
3) compute the \textbf{entropy} of such a distribution to measure \emph{spread},  
4) compute the \textbf{margin} between the top-1 and top-2 scores to measure \emph{decisiveness}, and  
5) combine these two metrics into a unified anomaly score.  
If the anomaly score exceeds a threshold, the query $q$ is flagged as malicious.

\textbf{Step 1. Top-$k$ Similarity Extraction.}  
Given a query embedding $q \in \mathbb{R}^d$ and record embeddings $\{r_i\}_{i=1}^n \subset \mathbb{R}^d$, we compute similarity scores $s_i = \langle q, r_i \rangle$ and sort them in descending order as $s_{(1)} \geq s_{(2)} \geq \cdots \geq s_{(n)}$. The top-$k$ scores $\{s_{(1)}, \dots, s_{(k)}\}$ are retained for analysis.  

\textbf{Step 2. Softmax Normalization.}  
To obtain a probabilistic interpretation, we normalize the top-$k$ scores using softmax: $p_i = \exp(s_{(i)}) / \sum_{j=1}^k \exp(s_{(j)})$, ensuring $\sum_{i=1}^k p_i = 1$, where $p_i$ reflects the retrieval weight of record $r_{(i)}$.  

\textbf{Step 3. Entropy (Retrieval Spread).}  
The distributional spread is measured via Shannon entropy $H = -\sum_{i=1}^k p_i \log p_i$, normalized as $\hat{H} = H / \log k \in [0, 1]$. A low $\hat{H}$ indicates abnormally concentrated retrieval, often symptomatic of malicious queries.  

\textbf{Step 4. Margin (Retrieval Decisiveness).}  
We compute the margin between the top-1 and top-2 score as $M = s_{(1)} - s_{(2)}$, and normalize it into $\hat{M} \in [0, 1]$ by min-max normalization to capture the decisiveness of the top match. Large $\hat{M}$ values indicate unusually dominant retrieval.  

\textbf{Step 5. Anomaly Detection.}  
Finally, we combine entropy and margin into a unified anomaly score using $A = \alpha \cdot (1 - \hat{H}) + \beta \cdot \hat{M}$, where $\alpha, \beta \geq 0$ are weighting coefficients. A query is flagged as malicious if $A \geq \tau$, with $\tau$ as the detection threshold.

\section{Evaluation}
\label{sec:evaluation}








\bheading{Scenarios.} 
We evaluate our defense in two real-world domains: \textbf{healthcare} and \textbf{personal assistance}. Specifically, we use the \textit{HealthcareMagic-101} dataset~\cite{li2023chatdoctor} (denoted as \texttt{Healthcare}), which contains patient-doctor dialogues, and the \textit{Enron Email} dataset~\cite{klimt2004enron} (denoted as \texttt{Email}).

\noindent\bheading{RAG components.}
We implement our RAG system using the LangChain framework, which modularizes the retrieval and generation pipelines. Documents are segmented into chunks and embedded using \texttt{all-MiniLM-L6-v2} for the \texttt{Healthcare} dataset and \texttt{BGE v1.5-large} for the \texttt{Email} dataset. For each query, the top-$k$ most relevant chunks ($k=3$) are retrieved and concatenated with the query to form the final prompt for generation.
We evaluate four representative LLMs: \texttt{Llama2-7b-chat}, \texttt{Llama2-13b-chat}, \texttt{Qwen2-72B-Instruct}, and \texttt{ChatGPT-4}.

\noindent\bheading{Metrics.}  
We use three metrics to quantify privacy leakage: \texttt{Unique Leakage Chunks (ULC)}, which measures the number of non-redundant document chunks containing private information; \texttt{PII}, the total number of individual PII elements leaked; \texttt{Attack Success Rate (ASR)} is defined as the fraction of adversarial queries that leak at least one PII item, with values ranging from 0 to 1.

\noindent\bheading{Attacks.}  
We evaluated recent  attacks, including prompt-injection methods TGTB~\cite{zeng2024good} and PIDE~\cite{qi2024follow}, as well as query-optimization attacks RAG-Thief~\cite{pal2020activethief} and Pirate~\cite{di2024pirates}.

\noindent\bheading{Baselines.}  
We compare against four categories of existing defenses: \emph{self-processing} (query rewriting~\cite{ma2023query}), \emph{auxiliary filtering} (summary- or rule-based filters~\cite{rahman2025summary}), \emph{summarization} (post-retrieval content reduction via LLMs), and \emph{erase-and-check}~\cite{kumar2023certifying} detecting adversarial prompts by iteratively truncating inputs with a safety filter.

\begin{table}[!ht]
    \centering
    \footnotesize
    \begin{tabular}{l|ccc|ccc}
    \hline
    \textbf{Model} 
        & \multicolumn{3}{c|}{\texttt{Healthcare}} 
        & \multicolumn{3}{c}{\texttt{Email}} 
        \\
    \cline{2-7}
    & \textbf{ULC} & \textbf{PII} & \textbf{ASR} 
        & \textbf{ULC} & \textbf{PII} & \textbf{ASR}  \\
    \hline 
 
     TGTB    & 105 & 78 & 0.75  & 112 & 117 & 1.00  \\
   PIDE    & 124 & 73 & 0.69  & 133 & 109 & 1.00 \\
    RAG-Thief        & 156 & 81 & 0.72  & 182 & 156 & 1.00   \\
     Pirate        & 212 & 97 & 0.93  & 238 & 201 & 1.00  \\
    \hline
    \end{tabular}
    \caption{Attack results without defense (100 prompts).}
    \label{tab:target-attack-nodefense}
\end{table}
    
\begin{table*}[!ht]
    \centering
    \resizebox{0.85\textwidth}{!}{
    \begin{tabular}{l|l|ccc|ccc|ccc|ccc}
    \hline
    \textbf{Defense} & \textbf{Model} 
        & \multicolumn{3}{c|}{\textbf{TBTG}} 
        & \multicolumn{3}{c|}{\textbf{PIDE}} 
        & \multicolumn{3}{c|}{\textbf{RAG-Thief}} 
        & \multicolumn{3}{c}{\textbf{Pirate}} \\   
    \cline{3-14}
     & & \textbf{ULC} & \textbf{PII} & \textbf{ASR} 
        & \textbf{ULC} & \textbf{PII} & \textbf{ASR} 
      & \textbf{ULC} & \textbf{PII} & \textbf{ASR} 
     & \textbf{ULC} & \textbf{PII} & \textbf{ASR} \\
    \hline 
    \multirow{4}{*}{{Rephrasing}} 
        & Llama2-7b-chat   & 98 & 75 & 0.71 & 109  & 48 & 0.48 & 119 & 52 & 0.43 & 199 & 100 & 1.00  \\
        & Qwen2-72B        & 101 & 65 & 0.60 &   120 & 67  & 0.67 & 138  & 56 & 0.35 & 203 & 81 & 0.75  \\
        & ChatGPT-4        & 112 &   81 & 0.76 & 129  & 50 & 0.50 & 149 & 85 & 0.70 & 208 & 94 & 0.90  \\
    \hline
    \multirow{4}{*}{{Summarization}} 
            & Llama2-7b-chat   & 87  & 60 & 0.52 & 102  & 55 & 0.45 & 95 & 63 & 0.55 & 176 & 82 & 0.72  \\
            & Qwen2-72B        & 94  & 66 & 0.60 & 111  & 59 & 0.53 & 124 & 71 & 0.61 & 182 & 90 & 0.85  \\
            & ChatGPT-4        & 98 & 74 & 0.68 & 106  & 62 & 0.40 & 135 & 78 & 0.68 & 191 & 96 & 0.87  \\
        \hline
    \multirow{4}{*}{{Rule-based}} 
       & Llama2-7b-chat   & 12 & 3 & 0.03 & 71  & 36 & 0.36 & 80 & 45 & 0.45 & 132 & 38 & 0.29  \\
        & Qwen2-72B        & 13 & 5 & 0.05 & 58  & 37 & 0.37 & 82 & 33 & 0.33 & 157 & 59 & 0.55  \\
        & ChatGPT-4        & 20 & 8 & 0.08 & 65  & 30 & 0.30 & 83 & 34 & 0.34 & 116 & 58 & 0.40  \\
    \hline
    \multirow{4}{*}{{Erase-check}} 
        & Llama2-7b-chat   & 35 & 12 & 0.12 & 13  & 3 & 0.03 & 66 & 15 & 0.15 & 97 & 60 & 0.47 \\
        & Qwen2-72B        & 43 & 17 & 0.17  & 25  & 10 & 0.10 & 74 & 21 & 0.21 & 120 & 64 & 0.53  \\
        & ChatGPT-4        & 52 & 19 & 0.19 & 22  & 7 & 0.07 & 72 & 27 & 0.27 & 127 & 71 & 0.68  \\
    \hline
    \multirow{4}{*}{\textbf{RAG-CT (Ours)}} 
        & Llama2-7b-chat   & \textbf{2} & \textbf{0} & \textbf{0.00} & \textbf{6}  & \textbf{0} & \textbf{0.00} & \textbf{22} & \textbf{3} & \textbf{0.03} & \textbf{67} & \textbf{11} & \textbf{0.08}  \\
        & Qwen2-72B        & \textbf{1} & \textbf{0} & \textbf{0.00} & \textbf{5}  & \textbf{0} & \textbf{0.00} & \textbf{24} & \textbf{5} & \textbf{0.05} & \textbf{63} & \textbf{9} & \textbf{0.05}  \\
        & ChatGPT-4        & \textbf{4} & \textbf{0} & \textbf{0.00} & \textbf{14}  & \textbf{0} & \textbf{0.00} & \textbf{26} & \textbf{4} & \textbf{0.04} & \textbf{50} & \textbf{6} & \textbf{0.06}  \\
    \hline
    \end{tabular}
    }
    \caption{Comparison of defense performance across attacks (\texttt{Healthcare} dataset, 100 prompts).}
    \label{tab:target-attack-datasets}
\end{table*}

\begin{figure}[t]
    \centering
    \begin{subfigure}{0.49\linewidth}
        \centering
        \includegraphics[width=\linewidth]{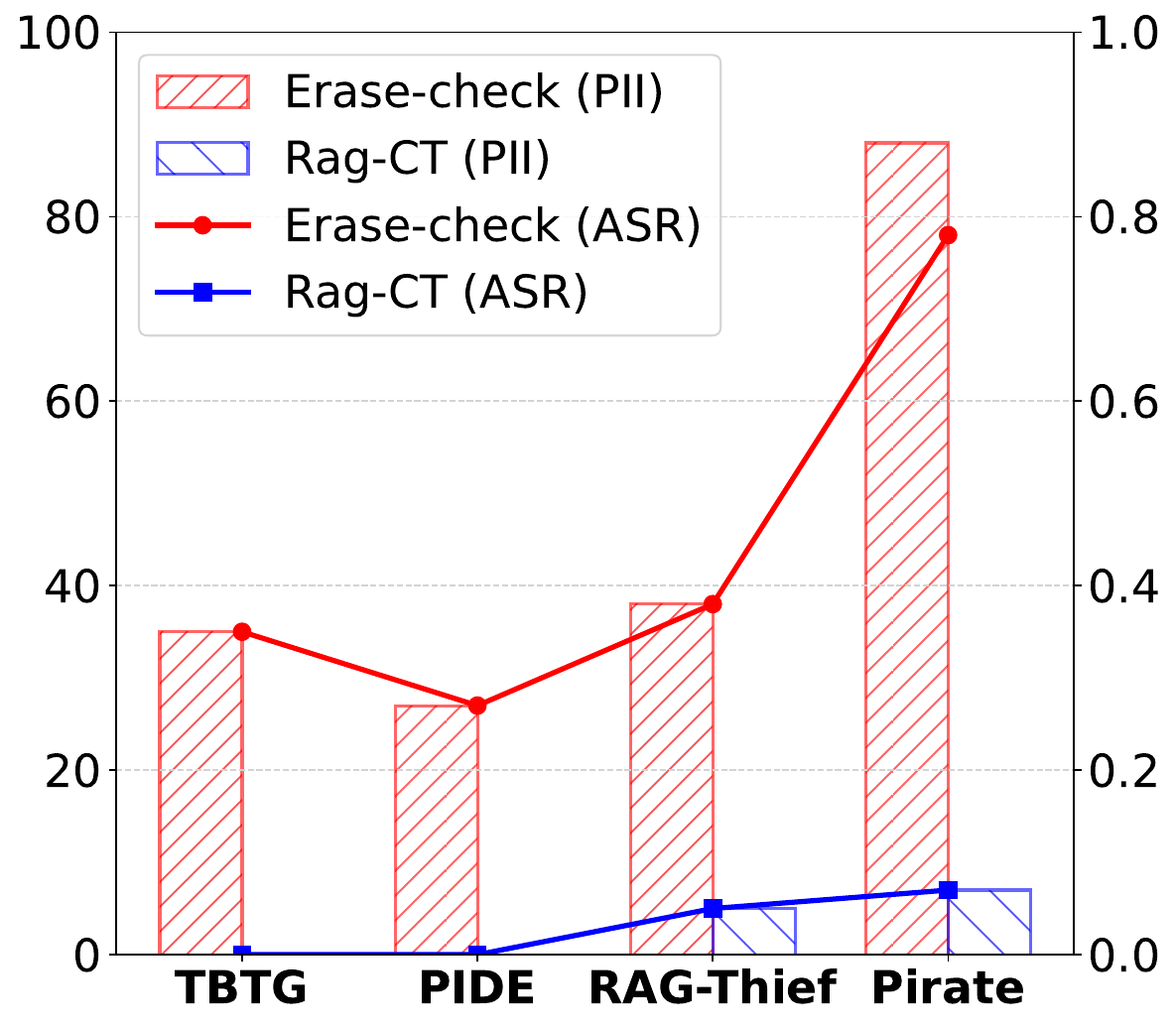}
        \caption{Defense on Email dataset}
        \label{fig:pii-asr-email}
    \end{subfigure}
    \hfill
    \begin{subfigure}{0.49\linewidth}
        \centering
        \includegraphics[width=\linewidth]{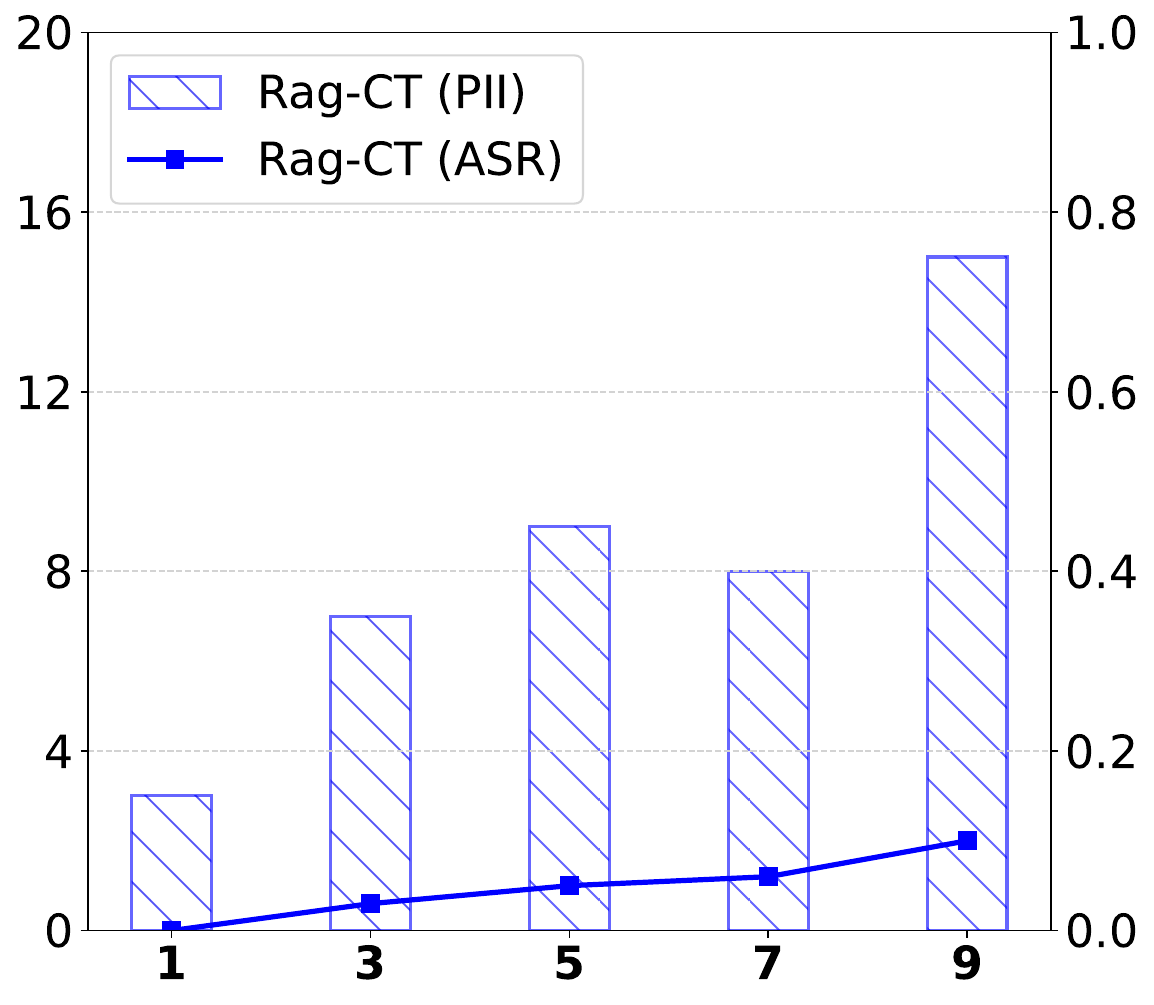}
        \caption{Top-k retrieval size}
        \label{fig:ablation-topk}
    \end{subfigure}
    \hfill
    \begin{subfigure}{0.49\linewidth}
        \centering
        \includegraphics[width=\linewidth]{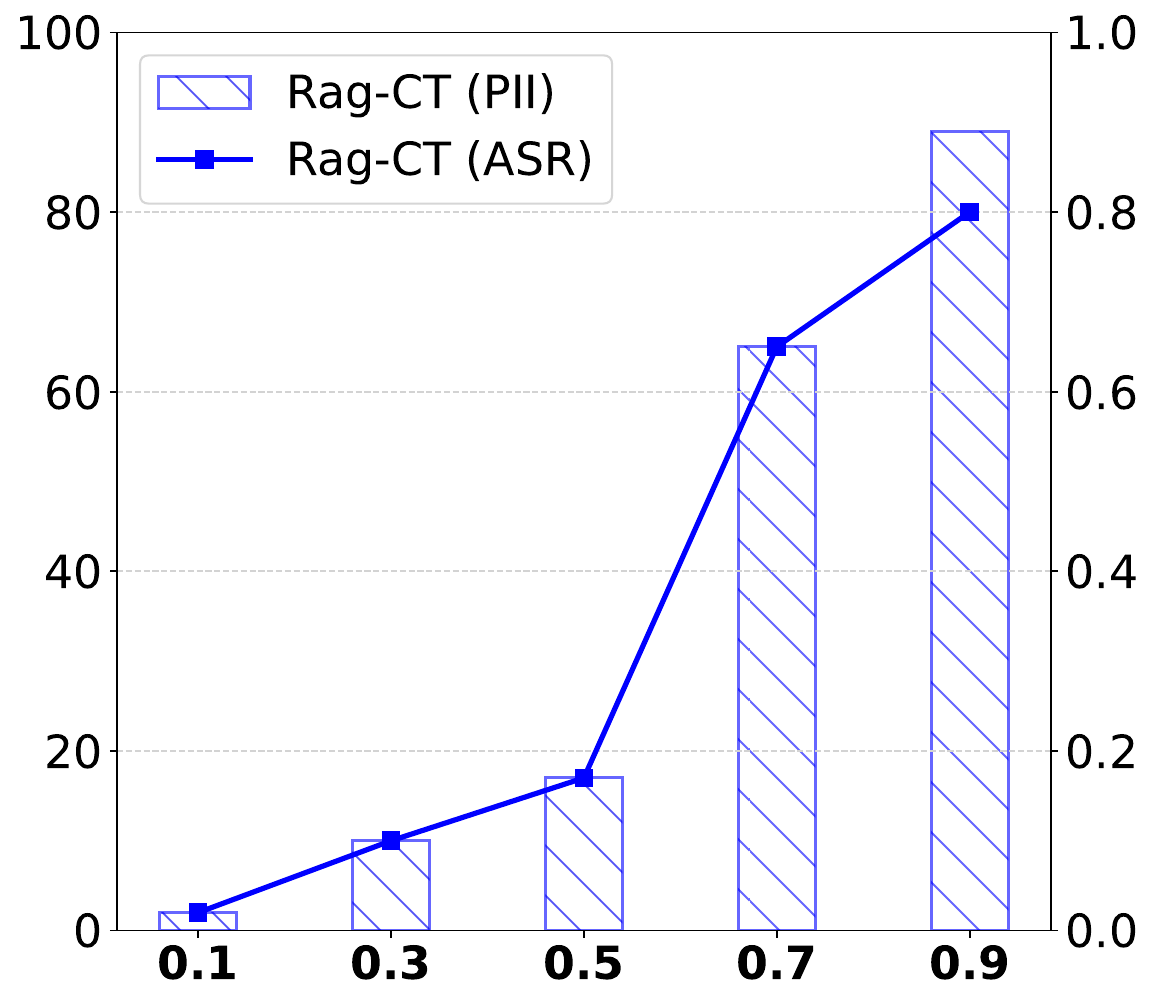}
        \caption{Detection threshold}
        \label{fig:ablation-threshold}
    \end{subfigure}
   \hfill
    \begin{subfigure}{0.49\linewidth}
        \centering
        \includegraphics[width=\linewidth]{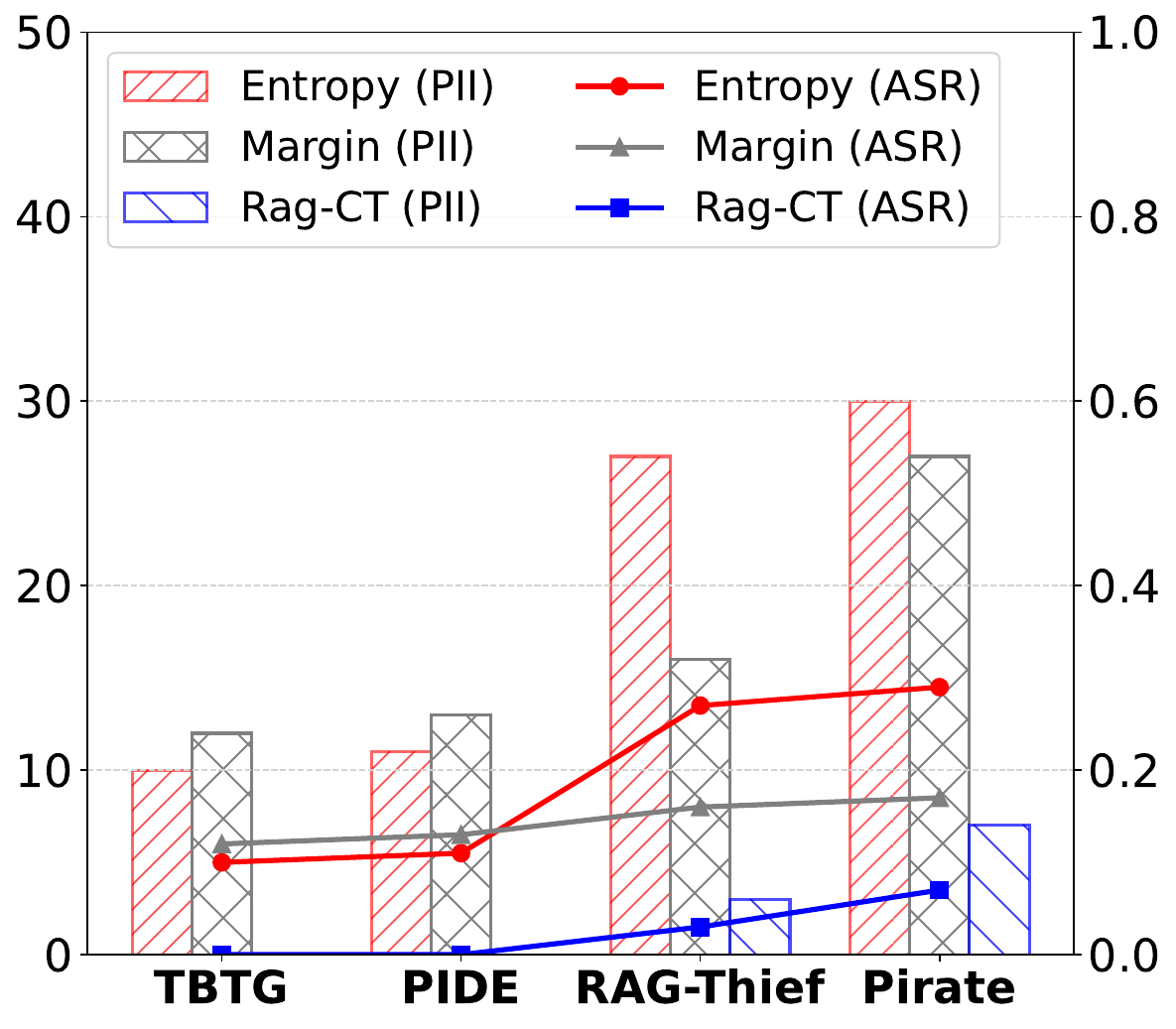}
        \caption{Detection strategy}
        \label{fig:ablation-intuition}
    \end{subfigure}
    \hfill

    \caption{(a): Defense results; (b)-(d): Ablation studies.}
    \label{fig:ablation-all}
\end{figure}

\noindent\bheading{Main results.}  
Table~\ref{tab:target-attack-nodefense} reports attack performance on \texttt{Llama2-8b-chat} with the \texttt{Healthcare} dataset using 100 prompts. In the absence of defenses, adversaries achieve high levels of PII leakage. In our evaluation, PII includes full name, phone number, email address, ID card number, SSN, passport, driver’s license, date of birth, and residential or work address.  
Table~\ref{tab:target-attack-datasets} summarizes results under five defense strategies on the \texttt{Healthcare} dataset. Rephrasing and summarization introduce only marginal reductions in leakage. Rule-based filtering is effective against TBTG and partially mitigates PIDE and RAG-Thief, but has limited impact on Pirate. Similarly, erase-and-check is effective against TBTG and PIDE but performs poorly against other attacks. \textbf{In contrast, our defense achieves consistent robustness} across all attacks: for TBTG and PIDE, the ASR is reduced to $0.00$ across all models and remains below $0.1$ even under RAG-Thief and Pirate. Similar trends are observed on the \texttt{Email} dataset, as shown in Fig.~\ref{fig:pii-asr-email}.

\noindent\bheading{Ablation study}.  
Fig.~\ref{fig:ablation-topk} shows our defense result under varying top-$k$ retrieval sizes  using \texttt{Llama2-7b-chat}, the Pirate attack, and the \texttt{Healthcare} dataset. It shows that retrieving more content increases the ASR. Fig.~\ref{fig:ablation-threshold} further illustrates that a lower detection threshold enforces stricter filtering, leading to stronger defense; we set the default threshold to $0.5$. Fig.~\ref{fig:ablation-intuition} compares three strategies: entropy, margin, and our proposed RAG-CT. We find that RAG-CT achieves the best performance, confirming the effectiveness of the proposed intuition in Sect.~\ref{intuition}.



\section{Related Work}
\label{sec:related}


\bheading{Privacy leakage on RAG systems.}  
Early prompt-injection attacks used static queries to extract limited private fragments~\cite{zeng2024good,cohen2024unleashing,qi2024follow}. Recent work advances to adaptive query-refinement, where queries are iteratively modified to uncover sensitive data. For example, Jiang et al.~\cite{jiang2024rag} introduce an agent-based approach with segment extensions, while Di Maio et al.~\cite{di2024pirates} design a relevance-driven mechanism that dynamically updates anchors and queries.  

\noindent\bheading{Defenses against privacy leakage in LLM/RAG.}  
Existing defenses fall into three main categories~\cite{xu2024comprehensive}: \emph{self-processing}, such as query rewriting~\cite{ma2023query}; \emph{auxiliary filtering}, including keyword- and semantic-based blocking~\cite{rahman2025summary}; and \emph{input permutation}, such as erase-and-check~\cite{kumar2023certifying}. Beyond these, cryptographic frameworks integrate differential privacy, secure multi-party computation, and homomorphic encryption to mitigate leakage~\cite{kandula2025securing}.

\section{Conclusion}
\label{sec:conclude}

In this paper, we proposed \textbf{RAG-CT}, a distribution-based defense against PII extraction attacks over RAG systems. By combining entropy and margin analysis, RAG-CT effectively detects and blocks queries targeting sensitive records. Extensive experiments demonstrate that our method outperforms state-of-the-art defenses. This study sheds light on defending RAG against privacy leakage and provides a lightweight and effective approach for safeguarding RAG systems.

\bibliographystyle{IEEEbib}
\bibliography{paper}

@article{li2023chatdoctor,
  title={Chatdoctor: A medical chat model fine-tuned on a large language model meta-ai (llama) using medical domain knowledge},
  author={Li, Yunxiang and Li, Zihan and Zhang, Kai and Dan, Ruilong and Jiang, Steve and Zhang, You},
  journal={Cureus},
  volume={15},
  number={6},
  year={2023},
  publisher={Cureus}
}

@article{latif2025hallucinations,
  title={Hallucinations in Large Language Models and Their Influence on Legal Reasoning: Examining the Risks of AI-Generated Factual Inaccuracies in Judicial Processes},
  author={Latif, Youssef Abdel},
  journal={Journal of Computational Intelligence, Machine Reasoning, and Decision-Making},
  volume={10},
  number={2},
  pages={10--20},
  year={2025}
}

@article{zeng2024good,
  title={The good and the bad: Exploring privacy issues in retrieval-augmented generation (rag)},
  author={Zeng, Shenglai and Zhang, Jiankun and He, Pengfei and Xing, Yue and Liu, Yiding and Xu, Han and Ren, Jie and Wang, Shuaiqiang and Yin, Dawei and Chang, Yi and others},
  journal={arXiv preprint arXiv:2402.16893},
  year={2024}
}

@article{qi2024follow,
  title={Follow My Instruction and Spill the Beans: Scalable Data Extraction from Retrieval-Augmented Generation Systems},
  author={Qi, Zhenting and Zhang, Hanlin and Xing, Eric and Kakade, Sham and Lakkaraju, Himabindu},
  journal={arXiv preprint arXiv:2402.17840},
  year={2024}
}

@article{cohen2024unleashing,
  title={Unleashing worms and extracting data: Escalating the outcome of attacks against rag-based inference in scale and severity using jailbreaking},
  author={Cohen, Stav and Bitton, Ron and Nassi, Ben},
  journal={arXiv preprint arXiv:2409.08045},
  year={2024}
}

@article{jiang2024rag,
  title={Rag-thief: Scalable extraction of private data from retrieval-augmented generation applications with agent-based attacks},
  author={Jiang, Changyue and Pan, Xudong and Hong, Geng and Bao, Chenfu and Yang, Min},
  journal={arXiv preprint arXiv:2411.14110},
  year={2024}
}

@article{di2024pirates,
  title={Pirates of the RAG: Adaptively Attacking LLMs to Leak Knowledge Bases},
  author={Di Maio, Christian and Cosci, Cristian and Maggini, Marco and Poggioni, Valentina and Melacci, Stefano},
  journal={arXiv preprint arXiv:2412.18295},
  year={2024}
}

@inproceedings{pal2020activethief,
  title={Activethief: Model extraction using active learning and unannotated public data},
  author={Pal, Soham and Gupta, Yash and Shukla, Aditya and Kanade, Aditya and Shevade, Shirish and Ganapathy, Vinod},
  booktitle={AAAI},
  volume={34},
  number={01},
  pages={865--872},
  year={2020}
}

@article{ram2023context,
  title={In-context retrieval-augmented language models},
  author={Ram, Ori and Levine, Yoav and Dalmedigos, Itay and Muhlgay, Dor and Shashua, Amnon and Leyton-Brown, Kevin and Shoham, Yoav},
  journal={Transactions of the Association for Computational Linguistics},
  volume={11},
  pages={1316--1331},
  year={2023},
  publisher={MIT Press One Broadway, 12th Floor, Cambridge, Massachusetts 02142, USA~…}
}

@inproceedings{klimt2004enron,
  title={The enron corpus: A new dataset for email classification research},
  author={Klimt, Bryan and Yang, Yiming},
  booktitle={European conference on machine learning},
  pages={217--226},
  year={2004},
  organization={Springer}
}

@misc{all-MiniLM-L6-v2,
  title = {all-MiniLM-L6-v2},
  howpublished = {\url{https://huggingface.co/sentence-transformers/all-MiniLM-L6-v2}},
  note = {Accessed: 2025-03-04}
}

@article{hindi2025enhancing,
  title={Enhancing the Precision and Interpretability of Retrieval-Augmented Generation (RAG) in Legal Technology: A Survey},
  author={Hindi, Mahd and Mohammed, Linda and Maaz, Ommama and Alwarafy, Abdulmalik},
  journal={IEEE Access},
  year={2025},
  publisher={IEEE}
}

@inproceedings{zhang2023enhancing,
  title={Enhancing financial sentiment analysis via retrieval augmented large language models},
  author={Zhang, Boyu and Yang, Hongyang and Zhou, Tianyu and Ali Babar, Muhammad and Liu, Xiao-Yang},
  booktitle={Proceedings of the fourth ACM international conference on AI in finance},
  pages={349--356},
  year={2023}
}

@misc{nvidia-ai-virtual-assistant,
  author       = {{NVIDIA}},
  title        = {{ai-virtual-assistant}: NVIDIA AI Virtual Assistant Blueprint},
  year         = {2024}
}

@manual{Amazon2025rag,
  author       = {{Amazon}},
  title        = {Creating Retrieval Augmented Generation solutions on AWS for healthcare},
  year         = {2025},
  organization = {Amazon Web Services},
  howpublished = {\url{https://docs.aws.amazon.com/prescriptive-guidance/latest/rag-healthcare-use-cases/introduction.html}}
}

@article{kandula2025securing,
  title={Securing Retrieval-Augmented Generation-Privacy Risks and Mitigation Strategies},
  author={Kandula, Sheshananda Reddy},
  journal={Available at SSRN 5191687},
  year={2025}
}

@article{xu2024comprehensive,
  title={A comprehensive study of jailbreak attack versus defense for large language models},
  author={Xu, Zihao and Liu, Yi and Deng, Gelei and Li, Yuekang and Picek, Stjepan},
  journal={arXiv preprint arXiv:2402.13457},
  year={2024}
}

@inproceedings{ma2023query,
  title={Query rewriting in retrieval-augmented large language models},
  author={Ma, Xinbei and Gong, Yeyun and He, Pengcheng and Zhao, Hai and Duan, Nan},
  booktitle={Proceedings of the 2023 Conference on Empirical Methods in Natural Language Processing},
  pages={5303--5315},
  year={2023}
}

@inproceedings{rahman2025summary,
  title={Summary the Savior: Harmful Keyword and Query-based Summarization for LLM Jailbreak Defense},
  author={Rahman, Shagoto and Harris, Ian},
  booktitle={TrustNLP 2025},
  pages={266--275},
  year={2025}
}

@article{kumar2023certifying,
  title={Certifying llm safety against adversarial prompting},
  author={Kumar, Aounon and Agarwal, Chirag and Srinivas, Suraj and Li, Aaron Jiaxun and Feizi, Soheil and Lakkaraju, Himabindu},
  journal={arXiv preprint arXiv:2309.02705},
  year={2023}
}

@article{choi2025safeguarding,
  title={Safeguarding Privacy of Retrieval Data against Membership Inference Attacks: Is This Query Too Close to Home?},
  author={Choi, Yujin and Park, Youngjoo and Byun, Junyoung and Lee, Jaewook and Park, Jinseong},
  journal={arXiv preprint arXiv:2505.22061},
  year={2025}
}

@INPROCEEDINGS{10884425,
  author={Tamang, Lakpa and Bouadjenek, Mohamed Reda and Dazeley, Richard and Aryal, Sunil},
  booktitle={ICDM}, 
  title={Margin-Bounded Confidence Scores for Out-of-Distribution Detection}, 
  year={2024},
  volume={},
  number={},
  pages={1-10},
  doi={10.1109/ICDM59182.2024.00053}}

\end{document}